\documentclass[conference,letterpaper]{_sty/IEEEtran}
\IEEEoverridecommandlockouts

\usepackage{algorithmic}
\usepackage{amsmath,amssymb,amsfonts}
\usepackage{array}
\usepackage{booktabs}
\usepackage{caption}
\usepackage{color}
\usepackage{comment}
\usepackage[inline]{enumitem}
\usepackage{epsfig}
\usepackage{etoolbox}
\usepackage{fancybox}
\usepackage{float}
\usepackage{fixltx2e}
\usepackage{graphicx}
\usepackage[breaklinks,colorlinks]{hyperref}
\usepackage{mathrsfs}
\usepackage{multirow}
\usepackage{multicol}
\usepackage[square,numbers,sort,compress]{natbib} 
    \def\BibTeX{{\rm B\kern-.05em{\sc i\kern-.025em b}\kern-.08em
    T\kern-.1667em\lower.7ex\hbox{E}\kern-.125emX}}
\usepackage{placeins}
\usepackage{rotating}
\usepackage{setspace}
\usepackage[hang, tight]{subfigure}
\usepackage{textcomp}
\usepackage{threeparttable}
\usepackage{times}
\usepackage{url}
\usepackage{verbatim}
\usepackage{wrapfig}
\usepackage[usenames,dvipsnames]{xcolor}
\usepackage[normalem]{ulem}

\usepackage[capitalize]{cleveref}
\crefname{section}{Sec.}{Secs.}
\Crefname{section}{Section}{Sections}
\Crefname{table}{Table}{Tables}
\crefname{table}{Tab.}{Tabs.}

\newbool{inccomment}
\setbool{inccomment}{true}
\newcommand{\XC}[1]{\ifbool{inccomment}{{\color{blue}XC\@: #1}}{}}
\newcommand{\JX}[1]{\ifbool{inccomment}{{\color{red}JX\@: #1}}{}}
\newcommand{\CX}[1]{\ifbool{inccomment}{{\color{magenta}CX\@: #1}}{}}
\newcommand{\TD}[1]{\ifbool{inccomment}{{\color{orange}#1}}{}}
\newcommand{\FN}[1]{\ifbool{inccomment}{{\color{OliveGreen}#1}}{}}
\newcommand{\GR}[1]{\ifbool{inccomment}{{\color{Tan}#1}}{}}
\newcommand{\SL}{\ifbool{inccomment}{{\color{magenta}\\============================================\\}}}
\newcommand{\RF}{\ifbool{inccomment}{{\color{green}~[R]}}}

\newcommand{\roma}[1]{\uppercase\expandafter{\romannumeral #1\relax}}
\graphicspath{{_fig/}}

\begin{document}
\title{\Large \textbf{QuadraNet: Improving High-Order Neural Interaction Efficiency \\with Hardware-Aware Quadratic Neural Networks}\vspace{-3mm}}


\author{\IEEEauthorblockN{Chenhui Xu\textsuperscript{*}, Fuxun Yu\textsuperscript{*}, Zirui Xu\textsuperscript{*}, Chenchen Liu\textsuperscript{$\dagger$}, Jinjun Xiong\textsuperscript{$\ddagger$}, Xiang Chen\textsuperscript{$*$,\#}}
\textsuperscript{*}\textit{George Mason University}, \textsuperscript{$\dagger$}\textit{University of Maryland}, \textsuperscript{$\ddagger$}\textit{University at Buffalo}, \textsuperscript{\#}\textit{Peking University}\\

\{cxu21,fyu2,zxu21\}@gmu.edu, ccliu@umbc.edu, jinjun@buffalo.edu, xchen26@gmu.edu
\vspace{-5mm}}

\maketitle
\captionsetup{font={footnotesize}}
\begin{abstract}

Recent progress in computer vision-oriented neural network designs is mostly driven by capturing high-order neural interactions among inputs and features.
	And there emerged a variety of approaches to accomplish this, such as Transformers and its variants.
However, these interactions generate a large amount of intermediate state and/or strong data dependency, leading to considerable memory consumption and computing cost, and therefore compromising the overall runtime performance.
To address this challenge, we rethink the high-order interactive neural network design with a quadratic computing approach.
	Specifically, we propose \textit{QuadraNet} --- a comprehensive model design methodology from neuron reconstruction to structural block and eventually to the overall neural network implementation.
	Leveraging quadratic neurons' intrinsic high-order advantages and dedicated computation optimization schemes, \textit{QuadraNet} could effectively achieve optimal cognition and computation performance.
	Incorporating state-of-the-art hardware-aware neural architecture search and system integration techniques, \textit{QuadraNet} could also be well generalized in different hardware constraint settings and deployment scenarios.
	The experiment shows that \textit{QuadraNet} achieves up to 1.5$\times$ throughput, 30\% less memory footprint, and similar cognition performance, compared with the state-of-the-art high-order approaches.
\end{abstract}

\section{\textbf{Introduction}}

\begin{figure}[t]
\vspace{1mm}
	\centerline{\includegraphics[width=2.4in]{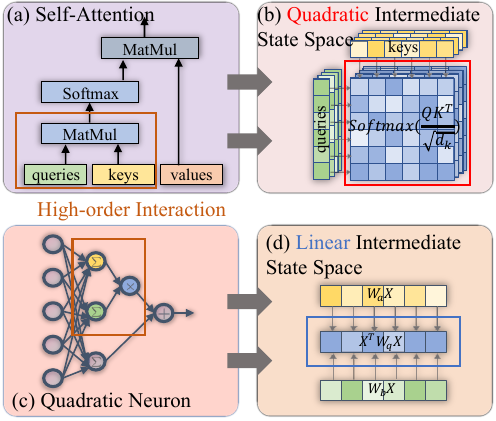}}
	 \caption{ High-Order Neural Interactions in Self-Attention Generations }
	\label{fig:problem}
	\vspace{-5mm}
\end{figure}

During the past decade, backbone neural networks for computer vision tasks have evolved from convolution-based information extraction~\cite{NIPS2012_c399862d,he2016deep} into Transformer-based semantic representation that well leverages self-attention mechanisms for information superposition~\cite{dosovitskiy2020image,liu2021swin}.
	However, such an advanced mechanism brings huge computation costs.
	Specifically, self-attention generation involves complex neural information interactions as shown in Fig.~\ref{fig:problem}~(a), resulting in more computing resource consumption and memory overheads.
	Therefore, even with supreme cognition performance, an extremely lightweight MobileViT model~\cite{mehta2021mobilevit} for edge devices still suffers a 10$\times$ latency slow-down than a CNN-based MobileNet\_V2~\cite{sandler2018mobilenetv2}, among which more than 60\% slow-down is caused by the resource-consuming self-attention generation.

As self-attention generation emerges as a new computing bottleneck for Transformer-like neural networks, an open question is how to better understand its in-depth working mechanism.
From the algorithmic perspective, self-attention generations are achieved by \textit{high-order neural interactions} of iterative point products across multiple transformed feature dimensions, generating input-adaptive weight parameters.
	Although such high-order neural interactions may best be captured at the cost of quadratic time and space complexity~\cite{keles2022computational}, a large number of sparse intermediate states are introduced as shown in Fig.~\ref{fig:problem}~(b).
	Compared to regular feature convolution operations with a linear complexity to the input size, these intermediate states consume a quadratic feature space, thus resulting the aforementioned high computation cost.
%
From the computing perspective, to avoid self-attention costs, some recent works proposed other means to achieve similar high-order neural interactions.
%
For example,~\cite{wang2020linformer} bypassed softmax operations to approximate self-attention as associative multiplication to avoid the quadratic intermediate states.~\cite{rao2022hornet} also devolved high-order production with gated recursive convolutions for lower computation complexity.
However, such approaches suffer from a number of problems. For example, many self-attention approximations fell into a sub-optimal feature space, compromising the cognition performance~\cite{wang2020linformer}; or some achieved comparable cognition performance but introduced more intricate data dependency, limiting not only the inference throughput but the model design generalization~\cite{rao2022hornet}.

Different from these state of arts~\cite{rao2022hornet,wang2020linformer}, in this paper, we propose to rethink the Transformer-like neural network constructions' in capturing the high-order neural interactions. Based on that understanding, we then propose a new neuron-level design coupled with neural architecture designs to address the high-order issues while retaining high accuracy, low computation, and great generalization capacities.
	Inspired by~\cite{chrysos2020p}, which builds neural networks with polynomials without activation functions, we found that high-order neural interactions could be directly embedded into high-order neuron designs.
	In particular, such neuron implementations could be achieved by \textit{quadratic neurons}~\cite{fan2018new,xu2022quadralib}.
	Quadratic neurons possess a strong ability for information self-reinforcement across multiple feature dimensions with  much simpler and more efficient computation patterns, therefore eliminating intermediate states from conventional self-attention generations.
	\textit{Thus, we take quadratic neurons as the key to the revolution of fundamental Transformer-like model designs.}

Following such an approach, in this paper, we establish a novel quadratic neural network architecture --- \textit{QuadraNet}, which masters high-order neural interactions and adapts to different hardware computing constraints.
	To take inspiration from quadratic neurons to a successful comprehensive \textit{QuadraNet} architecture realization, we need to tackle a series of challenges, including
	designing a more efficient quadratic neuron and leveraging emerging compiling techniques to accelerate quadratic convolution operations;
	building basic model structural blocks by incorporating quadratic convolutions into modern network models;
	and incorporating hardware awareness into the design of high-order interactive quadratic networks through a neural architecture search (NAS) approach.

The main contributions of this paper are as follows.
\begin{itemize}
	\item We demystify the high-order neural interactions in Transformer-like neural networks through comprehensive analyses of algorithms and computation patterns, and explore the bottleneck of existing optimization approaches.
	\item We break through conventional Transformer-like model designs with a full-stack innovation from quadratic neurons to structural blocks and eventually to comprehensive artificially designed and searched neural networks.
	\item We further generalize the quadratic neural networks design methodology to various model architectures and heterologous hardware computing constraints.
\end{itemize}

\textit{To the best of our knowledge, \textit{QuadraNet} is the very first quadratic neural network design methodology that achieves competitive cognition performance with state-of-the-art Transformer-like models while achieving outstanding computation efficiency.}
	Specifically, the experiment shows that the proposed \textit{QuadraNet} performs exceptional high-order neural interactions in terms of comparative self-attention representation capacity and classification accuracy.
	Meanwhile, \textit{QuadraNet} can provide 1.5$\times$ throughput without any accuracy reduction and 30\% memory occupation optimization compared with representative Transformer-like models~\cite{liu2021swin}.
\section{\textbf{Motivation Analysis}}

\subsection{\textbf{High-Order Interactive Neural Networks}}

\textbf{Existing Approaches:}
To capture high-order neural interactions, Transformer-like neural network models, such as ViT~\cite{dosovitskiy2020image} and its variants~\cite{liu2021swin}, adopt the self-attention mechanism. 
In conducting model cognition, self-attention mechanism performs a spatial mixing computation $y_{SA}^{(i,c)}$ of the $c^{th}$ output channel at the $i^{th}$ input feature dimension:
	\begin{equation}
		y_{Self-Attention}^{(i,c)} = \sum_{j \in \Omega_i}\sum_{c'=1}^CA(x)_{i\rightarrow j}T_V^{(c',c)}x^{(j,c')},
		\label{equ:self}
	\end{equation}
where $\Omega_i$ is the attention receptive field involving interactive neurons;
$A(x)$ is the input-adaptive weight matrix calculated by a softmax dot-production of ``query'' ($Q$) and ``key'' ($K$) transformations of the input $x$; and the $i$~$\rightarrow$~$j$ process denotes the relative position of the $i^{th}$ and $j^{th}$ feature dimension.
	Meanwhile, self-attention achieves channel-wise information transformation through its ``value'' ($V$) transformation of $T_V$ on $x^{(j,c')}$, and $(c',c)$ denotes the information transformation as manifested from the $c'^{th}$ input channel to the $c^{th}$ output channel.
	We argue that such a multiplication of $A(x)$ and $x$ forms high-order neural interactions, which is the key to such neural model's high cognition performance.

Different from the self-attention shown in Eq.~(\ref{equ:self}), HorNet~\cite{rao2022hornet} performs high-order neural interactions with a gated recursive convolution --- \textit{g$^r$Conv}, which can be formulated as:
\begin{equation}
	\hspace{-3mm}y_{g^rConv}^{(i,c)}=p_r^{(i,c)}=\sum_{j \in \Omega_i}\sum_{c'=1}^Cg(p_{r-1})w_{r-1,i\rightarrow j}^cT_{\phi_{in}}^{(c',c)}x^{(j,c')}.
	\label{equ:gnconv}
\end{equation}
Instead of generating high-order interactions directly from the input $x$, \textit{g$^r$Conv} uses a channel increasing recursive formulation $g(p_{r-1})$ to perform a high-order iteration of feature information, and thus getting an input-adaptive weight matrix. $T_{\phi_{in}}^{(c',c)}$ is a linear dimension switch input projection.

\textbf{Higher-Order Neural Interaction:}
By summarizing the above works, we conclusively find that both approaches adopt an input-dependent weight matrix and a channel-wise information transformation matrix.
	It can be formulated as a generic high-order neural interaction mechanism:
\begin{equation}
	y_{high-order}^{(i,c)} = \sum_{j \in \Omega_i}\sum_{c'=1}^Cg(x)_{ij}T^{(c',c)}x^{(j,c')},
	\label{equ:highorder}
\end{equation}
where $\Omega_i$ denotes the mechanism's receptive field; $g(x)$ denotes an interactive weight matrix that contains the input's information; and $T^{(c',c)}$ denotes a channel-wise information transformation.
	The multiplication of $x$ and input-adaptive $g(x)$ forms the high-order neural interaction of data itself.
	Such an interaction  leverages the superposition of activation information between different feature dimensions.

\subsection{\textbf{High-Order Computation Problems}}
The computation of existing high-order interaction approaches (esp., self-attention) face multiple problems regarding memory consumption and computing cost.

\textbf{Memory Consumption with Intermediate States:}
Conventional Transformer-like models usually suffer from excessive memory consumption. For example,
	in the training stage, all the data in intermediate states must be saved for backpropagations;
	in the inference stage, frequent and continuous generation and release of intermediate states during forward propagation also cast inconsistency and interval of memory operation and thus heavy IO consumption in runtime.
	Such consumption significantly restricts the neural model architecture's design flexibility and optimization feasibility.

As aforementioned, those intermediate states are introduced by self-attention generation.
	Specifically, the transformation process of $Q$, $K$, and $V$ parameters in Transformer-like models generates 3$HWC$ intermediate states, and the $softmax$($QK^T$)$V$ takes intermediate states as many as ($HW$)$^2$+($HW$)$^2C$, where $H$, $W$ and $C$ denote the height, width, and channels of the input per layer, respectively.
	Compared with widely-used depthwise separable convolution, which generates only $HWC$ intermediate states, self-attention needs $\sim$($HW$+3)$\times$ for computation.
	Even if the attention space is restricted to a local receptive field as in~\cite{liu2021swin}, the self-attention computation still yields $\sim$($M^2$+3)$\times$ more intermediate states, where $M$ denotes the field size (usually $\geq$7~\cite{liu2021swin}).

\textbf{Computing Cost with Data Dependency:}
In addition to memory consumption, the complex computing flow in self-attention also introduces huge costs.
	For example, the computation of \textit{g$^r$Conv} in HorNet makes use of recursive convolutions to achieve higher-order neural interactions.
	Although such ingenious design limits the FLOPs to an acceptable range, the recursive logic brings considerable data dependency issues.
	In other words, the next order of neural interactions in \textit{g$^r$Conv} must wait for the previous order's to complete.
	Therefore, a series of convolution operators are forced to be serialized, which largely compromises the computing parallelism performance and exacerbates computing latency, especially with GPUs.

Although such excessive intermediate states and data dependency bring new challenges to high-order interactive neural networks, the above motivation analysis also demystifies the core cognition mechanism of Transformer-like models and the computation problems.
	To achieve a revolutionary design with both cognition and computation performance in this work, we will show that quadratic neurons can play an important role.

\begin{figure}
	\centerline{\includegraphics[width=3.3in]{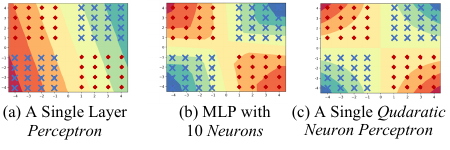}}
    \captionsetup{justification=centering}
    \vspace{-2mm}
    \caption{\footnotesize XOR Representation Capacity of Different Neurons \newline(colors: decision values, crosses/dots: positive/negative samples)}
    \vspace{-3mm}
	\label{fig:perceptron}
	
\end{figure}

\begin{figure*}[ht!]
	\centerline{\includegraphics[width=6.5in]{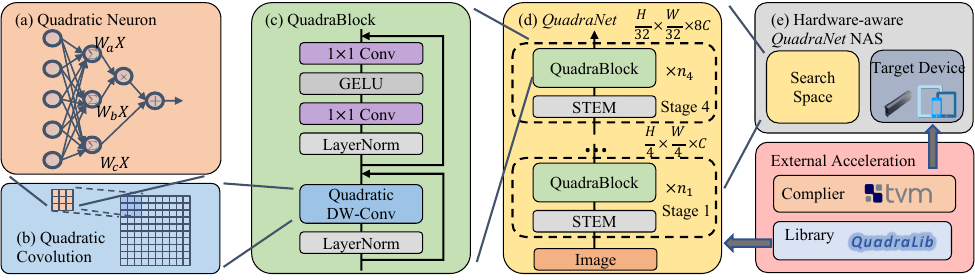}}
    \captionsetup{justification=centering}
	\caption{\footnotesize A Bottom-Up Overview of \textit{QuadraNet} (i.e., from neuron level to model level to system implementation level)}
	\label{fig:QuadraNet}
    \vspace{-3mm}
\end{figure*}

\subsection{\textbf{Quadratic Neuron}}
A quadratic neuron brings more opportunities to discover nonlinear relations among data dimensions because it performs a high-order interaction within the neuron for its quadratic computation.
	Therefore, with a shallower network, a quadratic neural network can learn a smoother and richer nonlinear representation than conventional neural networks.
	For example, for a generalized XOR classification problem, as shown in Fig.~\ref{fig:perceptron}, a single-layer perceptron with a quadratic neuron can solve it alone and gives even more precise classification boundaries than a two-layer perceptron with ten hidden neurons.

\textbf{Neuron Format:}
In contrast to the conventional one-order neuron represented by an inner product of input vector  $\mathbf{x} \in \mathbb{R}^{n \times 1}$ and weight vector $W \in \mathbb{R}^{1 \times n}$ plus a bias $b$ and activation, the quadratic neuron can be written as a quadratic polynomial of input $\mathbf{x}$ in the formulation:
	\begin{equation}
		\mathbf{y} = \mathbf{x}^TW_q\mathbf{x} + W_c\mathbf{x} + b,
		\label{equ:quadratic_neuron}
	\end{equation}
where $W_q$ and $W_c$ denote the quadratic- and linear-term weights, respectively.
	In Eq.~(\ref{equ:quadratic_neuron}), $\mathbf{x}^TW_q$ can be regarded as an input-dependent weight of $\mathbf{x}$; thus, it is evident that high-order interaction is directly embedded in a quadratic neuron~\cite{mantini2021cqnn,xu2022quadralib}.
        This refines the granularity of high-order interactions and their corresponding computation from the model architecture level to the neuron level, allowing more flexible adaption of high-order interactions.
	Therefore, we can leverage the neuron-level high-order interaction as described in Eq.~(\ref{equ:quadratic_neuron}) to build high-order neural networks with the mechanism in Eq.~(\ref{equ:highorder}).

\textbf{Opportunities and Challenges:}
By embedding high-order interactions into the neuron level, ordinary quadratic neurons also generate a quadratic parameter space as large as ($W_q \in \mathbb{R}^{n\times n}$), and its parameter volume rises to $O$($n^2$+$n$) compared with $O$($n$) with conventional neuron-based model implementations, since we now get an extra quadratic term weight matrix with $n^2$ learnable parameters.
    Subsequently, the computing workload in terms of Multiply Accumulate (MAC) operations rises to $O$($n^2$+$2n$).
	Therefore, an inappropriate implementation of quadratic neural networks would still cause certain computing issues, such as a significant parameter space dilation and thus an increase in the amount of computation.

This paper undertakes a meticulous investigation of the quadratic neural network, spanning from individual neurons to structural blocks, and ultimately to the overall model design and automatic search. The primary aim is to comprehensively address and optimize the parameter space dilation concern, while simultaneously preserving the quadratic neurons' robust high-order neural interaction capabilities.

\subsection{\textbf{Hardware-Aware Neural Architecture Search}}
Neural Architecture Search (NAS) has been extensively used in neural network design and optimization.
	Particularly, hardware-aware NAS considers computing cost, memory consumption, runtime latency, or power consumption as one of the optimization objectives in model architecture search algorithms, thus synthesizing a model that is well-balanced between cognition performance and computation efficiency.

Earlier on, little attention from the NAS area is given to high-order interactive neural networks due to the complex search space and large evaluation cost.
	Only very recently, the popularity of Transformer-like models triggered some works to build a NAS synthesized Vision Transformers~\cite{gong2021nasvit}.
	However, these works only focused on cognition performance without considering hardware awareness, not to mention the breakthrough in high-order neuron designs.

In this work, we target to incorporate hardware awareness into the proposed \textit{QuadraNet} design and expand its generalization capacity.
	Therefore, we will adopt mature NAS techniques into our design methodology to search for an adaptive high-order interactive neural network architecture for various deployment scenarios.
\section{\textbf{\textit{QuadraNet} Design Methodology}}
\subsection{\textbf{Design Methodology Overview}}

Fig.~\ref{fig:QuadraNet} illustrates the overview of our proposed methodology, which consists of the following major tasks:

(a) A novel computing-friendly \textit{quadratic neuron} design is proposed.
	With factorized low-rank quadratic terms, the parameter volumes and corresponding workload could be significantly optimized at the neuron level first.

(b) Based on quadratic neurons, a \textit{quadratic convolution} mechanism is built for high-order interactions.
	Different from conventional approaches (e.g., $g^rConv$), this mechanism could avoid large amounts of intermediate state spaces and recursive data dependency, further retaining the computation efficiency.

(c) \textit{QuadraBlock} is designed as the basic high-order neural network structural block.
	\textit{QuadraBlock} achieves high-order spatial neural interaction with quadratic depth-wise convolution and effectively unifies quadratic neurons within interactive block for optimal cognition capacity.

(d) A high-order neural network architecture \textit{QuadraNet} comprehensively integrates the proposed techniques above and leads a design revolution for Transformer-like models with outstanding cognition and computation performance.

(e) Hardware-aware NAS is further incorporated to improve the architecture generalization capacity of \textit{QuadraNet}, namely, the adaptability to various hardware computing constraints.

Eventually, external system implementation support is also well considered in this work, such as a full-stack compiler and library integration to target practical applications.
\subsection{\textbf{Efficient Quadratic Neuron and Convolution Design}}

\textbf{Quadratic Neuron:}
As defined in Eq.~(\ref{equ:quadratic_neuron}), the quadratic term of a conventional quadratic neuron has a parameter volume of $n^2$, which limits its usage in terms of both algorithm complexity and implementation difficulty.
	To address this issue, instead of using a full-rank quadratic weight matrix $W_q$, we propose to build the one-rank weight vectors $W_a$ and $W_b \in \mathbb{R}^{1\times n}$ as shown in Fig.~\ref{fig:QuadraNet} (a), which can be seen as a low-rank factorization of the quadratic weight term.
	Thus, the new quadratic neuron can be reformulated as:
\begin{equation}
	\mathbf{y} = \mathbf{x}^TW_a^TW_b\mathbf{x} + W_c\mathbf{x} + b\\
	=(W_a\mathbf{x})^T(W_b\mathbf{x}) + W_c\mathbf{x} + b.
	\label{equ:our_neuron}
\end{equation}

Such a factorization reduces the parameter space of quadratic neurons from $O$($n^2$+$n$) to $O(3n)$ and the computation complexity from $O$($n^2$+2$n$) to $O$(4$n$).

\textbf{High-Order Interactive Quadratic Convolution:}
Because quadratic neurons have powerful cognitive capacity and generalization ability, we can directly bring them into a simple way of information extraction.
	In this work, we directly project optimized quadratic neurons to the widely-used convolution operation to achieve efficient high-order interactions.
	Specifically, we map the three groups of weights $W_a$, $W_b$, and $W_c$ from a single quadratic neuron to three convolutional filters and conduct a polynomial computation like in Eq.~(\ref{equ:our_neuron}). With such a mapping, we obtain a quadratic convolution operator.
	Considering a mostly used depthwise separable convolution, the quadratic computation of the $i^{th}$ input data of the $c^{th}$ channel is formulated as follows:
\begin{equation}
	\begin{aligned}
		y_{Quad}^{(i,c)} &=\sum_{j,k \in \Omega_i} \sum_{c'=1}^C(\omega_{a,i\rightarrow j}T_{a}^{(c,c')}x^{(j,c')})(\omega_{b,i\rightarrow k}T_{b}^{(c,c')}x^{(k,c')})\\
		&=\sum_{j \in \Omega_i}\sum_{c'=1}^C\omega_{a,i\rightarrow j}\sum_{k \in \Omega_i}\omega_{b,i\rightarrow k}x^{(k,c')}T_{q}^{(c,c')}x^{(j,c')}\\
		&=\sum_{j \in \Omega_i}\sum_{c'=1}^Cq(x)_{ij}T_{q}^{(c,c')}x^{(j,c')},
	\end{aligned}
	\label{equ:q_conv}
\end{equation}
where $\Omega_i$ denotes the convolution receptive field in the $i^{th}$ position, and $q(x)_{ij}=\omega_{a,i\rightarrow j}\sum_{k \in \Omega_i}\omega_{b,i\rightarrow k}x^{(k,c')}$ implicates an input-adaptive weight.
When multiplying $x$, it captures the high-order spatial interaction.
	Notice that Eq.~(\ref{equ:q_conv}) forms a unified  expression for high-order interaction that well matches with Eq.~(\ref{equ:highorder}).
	This indicates that the quadratic convolution can capture high-order interactions, and therefore can be seen as an approximation computation of self-attention and \textit{g$^r$Conv}.

\textbf{Efficient High-Order Computation:}
Through the neuron- and convolution-level design, the primary high-order quadratic computation mechanism could be determined microscopically.
	From a tensor computation perspective, the computing of this mechanism could be formulated as follows:
\begin{equation}
	X_{out} = f_a(X_{in})\odot f_b(X_{in})+f_c(X_{in}),
	\label{equ:tensor}
\end{equation}
where $\odot$ denotes the Hadamard product; $f_a,f_b,f_c$ denote convolution operators (with weights of $W_a$, $W_b$, $W_c$); $X_{in}$ and $X_{out}$ denote the input and output tensors, respectively.
	Different from~\cite{rao2022hornet}, such a mechanism goes through a straightforward computation flow without a recursive logic, and thus eliminating the aforementioned data dependency and corresponding computing and memory costs.

As for the problem of intermediate states, we need to examine the ones generated in Eq.~(\ref{equ:tensor}).
	During forward propagation, four intermediate states are generated --- $f_a(X_{in})$,$f_b(X_{in})$,$f_a(X_{in})\odot f_b(X_{in})$, and $f_c(X_{in})$, all of which require the same data volume and 4$HWC$ intermediate state workload in total.
	The corresponding workload of intermediate states in a quadratic convolution will therefore be $4\times$ compared with a conventional convolution operator (i.e., $HWC$) for better cognition performance, but be much more efficient compared with the self-attention mechanism (i.e., ($HW$+3)$\times$) for less computing costs.

Meanwhile, during backpropagation, the $4\times$ intermediate states can be further reduced to $2\times$, since two intermediate states of $f_a(X_{in})\odot f_b(X_{in})$ and $f_c(X_{in})$ can be immediately released during computing before causing dominant workload.
	For the rest two states --- $f_a(X_{in})$ and $f_b(X_{in})$, weight updates of $W_a$ (or $W_b$) only rely on the intermediate state of $f_b(X_{in})$ (or $f_a(X_{in})$), because the gradient of the weight $W_a^l$ (or $W_b^l$) can be computed by:

\-\vspace{-3.5mm}
\footnotesize
\begin{equation}
	\begin{aligned}
	\frac{\partial \mathscr{L}}{\partial W_a^l} = &\frac{\partial \mathscr{L}}{\partial X^{l+1}}\cdot \frac{\partial X^{l+1}}{\partial (W_a^lX^l)(W_b^lX^l)}\cdot \frac{\partial (W_a^lX^l)(W_b^lX^l)}{\partial (W_a^lX^l)} \cdot \frac{\partial (W_a^lX^l)}{\partial W_a^l}\\
		= &\frac{\partial \mathscr{L}}{\partial X^{l+1}}\cdot (W_b^lX^l)\cdot X^l,
	\end{aligned}
\label{equ:garident}
\end{equation}
\normalsize
where $X^l$ denotes the $l^{th}$ layer output.
Moreover, the weight update of the linear term $W_c$ does not rely on any intermediate states.
	Therefore, only two intermediate states --- $f_a(X_{in})$ and $f_b(X_{in})$ are actually counted for memory and computing costs.

In a word, quadratic convolution enables more efficient higher-order computation by avoiding a recursive design for fewer data dependencies and further optimizations for intermediate states.
	Quadratic convolution also offers the possibility of external acceleration by simply mapping quadratic neurons to the convolution operators, allowing us to directly leverage existing compilers~\cite{chen2018tvm} and the quadratic computation library~\cite{xu2022quadralib} to optimize the computation.

\subsection{\textbf{Efficient and Powerful \textit{QuadraBlock}}}

\textbf{\textit{QuadraBlock} Design:}
Our QuadraBlock design is as shown in Fig.~\ref{fig:QuadraNet} (c).
	It is divided into a spatial feature mixing layer and an inter-channel information mixing layer:
	the spatial feature mixing layer is composed of a layer normalization operator and a quadratic depth-wise convolution;
	the inter-channel information mixing layer comprises a layer normalization operator, a 1$\times$1 convolution operator with a channel expansion coefficient $R$, a GELU activation, and another 1$\times$1 convolution operator restoring the number of channels.

Such a design achieves high computing efficiency by using only quadratic convolutions in spatial information mixing.
	Because in the modern CNNs design, the spatial feature mixing layers take less than 5\% of the parameters and MACs while performing important texture, outline, and shape depicting~\cite{liu2022convnet}.
	Therefore, inserting the quadratic convolutions into the spatial information mixture has little effect on the overall number of parameters and computation cost.

\textbf{\textit{QuadraBlock} for Cognition Capacity:}
	\textit{QuadraBlock} also unifies multiple quadratic neurons using channle-wise information transportation and activation. It enhances the cognition ability of quadratic neurons by approximating a high-rank quadratic weight matrix within the \textit{QuadraBlock}.
	This enhancement approximation is achieved by leveraging the nonlinear activation function to construct a linearly independent quadratic term weights matrix as in Eq.~(\ref{equ:rank}).
	The quadratic term of the combination of quadratic neuron layers, 1$\times$1 convolution with expansion coefficient $R$ and activation function $\sigma(\cdot)$ can be formulated as follows:
	\begin{equation}
		\begin{aligned}
			y_{comb}^c&=\sum_{c'=1}^{RC}\sigma(T_{q}^{(c,c')}\mathbf{x}^TW_a^TW_b\mathbf{x})\\
			&\approx \sum_{c'=1}^{C}T_{q}^{(c,c')}\mathbf{x}^T(\sum_{r=1}^{R}W_{ar}^TW_{br})\mathbf{x},
		\end{aligned}
		\label{equ:rank}
	\end{equation}
where $(\sum_{r=1}^{R}W_{ar}^TW_{br}) = W_A^TW_B, W_A,W_B \in \mathbb{R}^{n*R}$.

Thus, the delicately-designed \textit{QuadraBlock} enhances the representation capacity of quadratic weight matrix $W_q$ with $rank$=1 to a similar level as a $rank$=$R$, retaining both low-rank computation efficiency and cognition capability.

\subsection{\textbf{\textit{QuadraNet} Architecture Design:}}
Figure~\ref{fig:QuadraNet}~(d) illustrates the macro design of \textit{QuadraNet}.
We follow full-fledged four-stage pyramid architecture as \cite{liu2021swin,liu2022convnet}. which contains 4-stage with increasing channels and decreasing resolutions.
	In each stage, STEM~\cite{liu2022convnet} is first utilized for downsampling while increasing the number of channels, then the \textit{QuadraBlock} is repeated for $n_i$ times in stage i.
	And following configurations will be considered:

\begin{table}[t]
    \footnotesize
 \vspace{1.5mm}
	\caption{Cognition and Computation Performance Evaluation}
 \vspace{-1mm}
    \centering
    \tabcolsep 4.3pt
    \begin{tabular}{|l|ccc|cc|}

    \hline
        Model & \#Para. & FLOPs & \#Layer & Top-1 Acc. & Throughput \\ \
        ~ & (M) & (G) & ~ & (\%) & (img/s) \\ \hline
        Swin-T\cite{liu2021swin} & 29 & 4.5 & 12 & 81.3 & 1321 \\
        DeiT-S\cite{pmlr-v139-touvron21a} & 22 & 4.6 & 12 & 79.8 & 1621 \\
        TNT-S\cite{han2021transformer} & 23.8 & 5.2 & 12 & 81.5 & 657 \\
        PVT-S\cite{wang2021pyramid} & 24.5 & 3.8 & 16 & 79.8 & 1433 \\
        T2T-ViT-14\cite{yuan2021tokens} & 21.5 & 4.8 & 14 & 81.5 & 1376 \\
        ConvNeXt-T\cite{liu2022convnet} & 28 & 4.5 & 18 & 82.1 & 1944 \\
        HorNet-T\cite{rao2022hornet} & 22 & 4 & 25 & \textbf{82.8} & 1254 \\
        \textbf{QuadraNet25-T} & 16.2 & 2.9 & 25 & 81.2 & \textbf{2433} \\
        \textbf{QuadraNet36-T} & 23.6 & 4.1 & 36 & 82.2 & \textbf{1971} \\
        \hline

        Swin-S\cite{liu2021swin} & 50 & 8.7 & 24 & 83.0 & 827 \\
        ConvNeXt-S\cite{liu2022convnet} & 50 & 8.7 & 36 & 83.1 & 1275 \\
        HorNet-S\cite{rao2022hornet} & 50 & 8.8 & 25 & \textbf{83.8} & 813 \\
        \textbf{QuadraNet36-S} & 50.2 & 8.9 & 36 & \textbf{83.8} & \textbf{1117} \\ \hline

        Swin-B\cite{liu2021swin} & 88 & 15.4 & 24 & 83.5 & 662 \\
        ConvNeXt-B\cite{liu2022convnet} & 89 & 15.4 & 36 & 83.8 & 969 \\
        HorNet-B\cite{rao2022hornet} & 87 & 15.6 & 25 & \textbf{84.2} & 641 \\
        \textbf{QuadraNet25-B} & 61.2 & 11.1 & 25 & 83.6 & \textbf{1138} \\
        \textbf{QuadraNet36-B} & 90.4 & 15.8 & 36 & \textbf{84.1} & \textbf{892} \\ \hline
    \end{tabular}
    \label{tab:acc}
    \vspace{-3mm}
\end{table}

\textbf{Receptive field:}
We set the receptive field regarding convolution kernel size to 7$\times$7 for a quadratic depth-wise convolution.
	Compared with conventional 3$\times$3 practice, a large receptive field helps to capture high-order spatial interactions.

\textbf{Number of Channels:}
	In each stage, we set the number of channels in a gradually expanding way to [C, 2C, 4C, 8C] in each stage following common practice, specifically, C=[16, 32, 64, 96, 128] to construct  \textit{QuadraNet}~L-XXS/XS/T/S/B models (for different layers and scales).

\textbf{Number of Layers:}
	To fairly compare it with baselines~\cite{liu2021swin,liu2022convnet,rao2022hornet}, we can set the number of blocks as [3,3,27,3] and [2,3,18,2] in each stage to construct \textit{QuadraNet}~36/25-C with similar model volumes and number of layers.

\begin{table}[t]
    \footnotesize
    \vspace{1.5mm}
    \caption{Memory Consumption Evaluation}
    \vspace{-1mm}
    \centering
        \tabcolsep 5.3pt
    \begin{tabular}{|l|c|cc|}
    \hline
        Block &High-order& \multicolumn{2}{c|}{Traning Memory(MB)}  \\ 
        ~ & ~ & C=64 & C=128 \\ \hline
        SkipBlock(No DW-Conv) & $\times$ & 5515 & 12027 \\ 
        ConvBlock(Naive DW-Conv) & $\times$ & 7364 & 15711 \\ 
        SwinBlock\cite{liu2021swin} & \checkmark & 12674 & 23472 \\ 
        HorBlock\cite{rao2022hornet}  & \checkmark & 10724 & 20715 \\ 
        \textbf{QuadraBlock} & \checkmark & 8108 & 17191 \\ 
        \textbf{QuadraBlock*(w/ QuadraLib)}  & \checkmark & 7794 & 16730 \\ \hline
    \end{tabular}
    \label{tab:mem}
    \vspace{-3mm}
\end{table}

\subsection{\textbf{Hardware-Aware Quadratic Neural Architecture Search}}
We further incorporate state-of-the-art hardware-aware NAS techniques to further enhance \textit{QuadraNet}'s architecture generalization capability and hardware adaptability.

\textbf{Search Space:}
Following a common NAS development approach, \textit{QuadraNet} first requires a new parameter search space.
	When we generalize the above configuration options into automation search algorithm, we can therefore obtain a unique high-order interactive search space.
	Following the original ProxylessNAS~\cite{cai2018proxylessnas}, we specify a candidate set of \textit{QuadraBlock} with different kernel sizes \{3, 5, 7\} and expansion coefficients \{2, 4\}.
	We also add an identity block that does nothing but provides a residual~(skip) connection to the candidate set to search for a shallow network.

\textbf{Hardware Awareness:}
As demonstrated in previous sections, we have  analyzed the \textit{QuadraNet}'s memory and computing costs, as well as optimization factors.
	Given different systems' hardware profiling regarding different neural network operations, we can project different \textit{QuadraNet} model designs into particular resource consumption estimations.
	Therefore, the hardware-aware search feedback can be well established in the NAS process considering the hardware constraints.

\textbf{Search Method:}
Leveraging these efforts, we modify ProxylessNAS~\cite{cai2018proxylessnas} into a \textit{QuadraNet}-specified NAS framework by adapting the Search Space and feedback configuration. 
	Firstly, we train a supernet in defined \textit{QuadraNet} Search Space. Then with a gradient-based search strategy, considering hardware latency as a part of the loss function.
	And eventually, we train the branch weights to search for an appropriate architecture for the target device with particular hardware constraints.

\section{\textbf{Experiment}}
Since current high-order interactive neural networks are primarily tailored for image classification, we also apply the \textit{QuadraNet} methodology to this specific task.
	Specifically, we conduct experiments on ImageNet-1K dataset~\cite{deng2009imagenet}, which contains 1.28M training images and 50K validation images from 1000 classes.
	We report ImageNet-1K's Top-1 accuracy on the validation set.
	In the following, we compare the proposed \textit{QuadraNet} with state-of-the-art models regarding both cognition and computation performance.

\subsection{\textbf{\textit{QuadraNet} Evaluation with Manually Designed Models}}

\textbf{Model Initialization:}
We train \textit{QuadraNets} for 300 epochs (20 of which are warmup) with AdamW~\cite{loshchilov2018decoupled} optimizer and a learning rate of 4e-3.
	We use a weight decay of 0.05 and the same data augmentations as in~\cite{rao2022hornet}.
	The resolution of the input images is 224$\times$224.
	To stabilize the training, we apply gradient clipping that limits the maximum value of gradient to 5, instead of the superabundant batch normalization as in~\cite{xu2022quadralib}, which will hurt the generalization ability of networks.
	We use the ``Channel Last'' memory layout following~\cite{liu2022convnet}.

\textbf{Cognition Performance and Computing Costs:}
Table~\ref{tab:acc} evaluates the performance of several manually designed \textit{QuadraNet} models with other state-of-the-art models at comparable FLOPs and parameters.
 Our models demonstrate competitive performance across both Transformer-like and CNN-like architectures.
	With such a cognition performance baseline, we measure their inference throughputs on an A100 GPU\@.
	And results show that, with a similar image classification accuracy, \textit{QuadraNets} achieve 1.35$\times$ to 1.57$\times$ higher inference throughputs than high-order works~\cite{liu2021swin,rao2022hornet}.
Compared with ConvNeXt~\cite{liu2022convnet},
\textit{QuadraNets} achieve higher cognitive performance and even higher throughputs most of the time. 

\textbf{Memory Consumption Analysis:}
We further compare another set of models with a similar configuration except for different types of computation blocks to evaluate \textit{QuadraNet}'s memory consumption with 64 batch size, 24 layers, and 64/128 channels.
	Two low performant non-high-order models (SkipBlock and ConvBlock) are also involved in examining the trade-off between high-order interactive performance and costs.
	To make a fair comparison, the SkipBlock completely removes the spatial neural interaction and related residual connection, and the ConvBlock uses the same block design but only with the conventional depth-wise convolution.

Table~\ref{tab:mem} shows that, under the same configuration, \textit{QuadraBlock} achieves high-order information interaction with a relatively smaller extra memory footprint than SwinBlock and HorBlock.
	When coupled with accelerate library QuadraLib~\cite{xu2022quadralib} that further optimizes the intermediate state, \textit{QuadraBlock} results in an even smaller memory footprint.

\begin{table}[!b]
   \vspace{-3mm}
    \footnotesize
    \centering
    \caption{Hardware-Aware \textit{QuadraNet} NAS}
    \vspace{-1mm}
    \tabcolsep 4.4 pt
    \begin{tabular}{|l|ccccc|}
    \hline
        Searched Model & FLOPs & CPU & VPU & GPU & Top-1 Acc. \\ \hline
        Proxyless-CPU  & 333M & 299.6ms & 53.1ms & 1.8ms & 72.4 \\ 
        Proxyless-VPU  & 275M & 379.1ms & 29.9ms & 1.7ms & 72.7 \\ 
        Proxyless-GPU  & 1065M & 1521.1ms & 146.1ms & 2.9ms & 74.4 \\ \hline
        \textbf{QuadraNet\_CPU} & 279M & 299.8ms & 34.7ms & 2.1ms & 73.1 \\ 
         \textbf{QuadraNet\_VPU} & 312M & 437.1ms & 29.9ms & 1.9ms & 73.2 \\ 
         \textbf{QuadraNet\_GPU} & 948M & 1334.2ms & 117.5ms & 2.9ms & \textbf{76.7} \\ \hline
    \end{tabular}
    \label{tab:nas}
    
\end{table}
\subsection{\textbf{Hardware-Aware QuadraNet Model Generalization}}
\begin{figure}[t]

	\centering
	\includegraphics{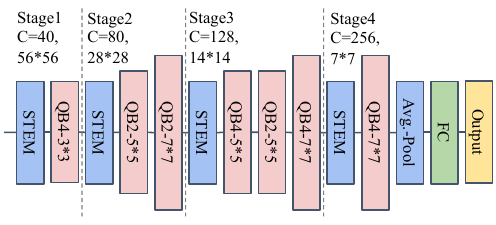}
 \vspace{-1mm}
	\caption{Visualization of Searched \textit{QuadraNet\_VPU\_10ms}}
	\label{fig:Nas}
\vspace{2mm}
\end{figure}

We conduct hardware-aware NAS with the help of nn-Meter~\cite{zhang2021nn} for its powerful latency estimation for different devices.
	We set the maximum latency for mobile CPU (CortexA76), VPU (Intel Myriad VPU), and GPU (A100) to 300ms, 30ms, and 3ms, respectively.
	Therefore, we compare the NAS models from regular ProxylessNAS~\cite{cai2018proxylessnas} and hardware-aware \textit{QuadraNet}.
	The results are shown in Table~\ref{tab:nas}.
	We observe that, given different hardware constraints, \textit{QuadraNet} models always perform better in terms of accuracy, FLOPs, and computation time.
	Fig~\ref{fig:Nas} also gives an example of a searched model with less than 10ms latency on VPU.

\begin{figure}[t]
	\vspace{-2mm}
	\centering
	\includegraphics{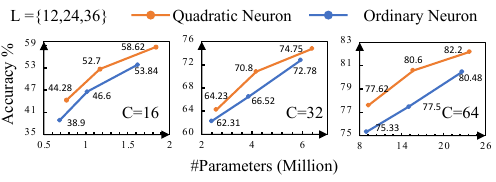}
     \vspace{-1mm}
	\caption{Outstanding Quadratic Performance w/ Various Configurations}
	\label{fig:my_label}
 \vspace{-3mm}
\end{figure}

\subsection{\textbf{Ablation Study}}

\textbf{Superiority of Quadratic Neuron:}
We further quantitatively study the cognitive capacity of quadratic neurons.
Fig.~\ref{fig:my_label} shows that, with different numbers of channels and layers, the additional parameters introduced by quadratic neurons can be well reflected in the consistent cognitive performance improvement, which further demonstrates the effectiveness of higher-order spatial neural interactions.

\textbf{Receptive Field's Impact:}
As high-order interaction is the essential mechanism of this work, we would take the receptive field as a representative factor to examine its correlation with \textit{QuadraNet} design.
    We train the \textit{QuadraNet36-T} with different kernel sizes on 4 A100 GPUs to investigate the impact of the receptive field.
As shown in Table~\ref{tab:receptive}, larger receptive fields indeed improve accuracy, but at the cost of considerable training time due to the inflated computational scale and unoptimized operators. 
    Therefore, this observation provides a new perspective on the cognition and computation trade-off.
    As mentioned above, we utilized a balance 7$\times$7 kernel, while more investigation would be performed in the future.

\textbf{Channel/Spatial-Wise Quadratic Neuron Utilization:}
After all, quadratic neurons have more computation costs than ordinary ones.
    Thus appropriate utilization of them becomes critical.
In conventional model designs, spatial-wise neurons only account for 5\% while channel-wise accounts for 95\%.
    When it comes to quadratic settings, a dedicated analysis is shown in Table~\ref{tab:fc} to identify which approach is more quadratic neuron-friendly.
    We can see that, although spatial-wise utilization is very small, it has even better effectiveness overall, highlighting a \textit{   QuadraNet} design principle.

\section{\textbf{Conclusion}}
In this work, we resented \textit{QuadraNet} --- a revolutionary neural network design methodology that incorporates efficient and powerful high-order neural interaction interactions.
	\textit{QuadraNet} avoids the existing high-order computational issues of intermediate states and data dependency down to the neuron design level.
	Based on the design of efficient quadratic neurons into \textit{QuadraBlock}, we construct a new family of generic backbone networks \textit{QuadraNets}, and provide a new search space for high-order interactions for hardware-aware NAS.
	Experiments demonstrate the superior cognitive capacity, high throughputs, and low training memory footprints of the proposed \textit{QuadraNet}.
	Hardware-aware NAS further identifies better quadratic neural network architecture under this new  search space with quadratic neurons.

\begin{table}[!t]
\footnotesize
\vspace{1.5mm}
    \centering
    \caption{Receptive Field's Impact}
    \vspace{-1mm}
    \tabcolsep 4.1pt
    \begin{tabular}{|l|cccccc|}
    \hline
        Receptive Field & 3*3 & 5*5 & 7*7 & 9*9 & 21*21 & Global Filter~\cite{rao2021global} \\ \hline
        Top-1 Acc. & 80.6 & 81.3 & 82.2 & 82.2 & 82.8 & 82.6 \\ 
        Training hours & 81.1 & 86.9 & 94.4 & 123.3 & 390.0 & 336.1 \\ \hline
    \end{tabular}
    \label{tab:receptive}
\end{table}
\begin{table}[!t]
    \footnotesize
    \centering
    \caption{
    Channel-wise Quadratic Operator's Impact
    }
    \vspace{-1mm}
    \tabcolsep 9.5pt
    \begin{tabular}{|l|ccc|}
    \hline
        Models  & \#Params & \#FLOPs & Top-1 Acc. \\ \hline
        QuadraNet36-T & 23.6M & 4.1G & 82.2 \\ 
        ~+Quadratic 1*1 Conv1 & 44.6M & 7.8G & 82.3 \\ 
        ~~+Quadratic 1*1 Conv2 & 65.6M & 11.5G & 82.7 \\ \hline
        QuadraNet36-B & 90.4M & 15.8G & 84.1 \\
        ~+Quadratic 1*1 Conv1 & 174.3M & 30.6G & 83.9 \\ 
        ~~+Quadratic 1*1 Conv2 & 258.2M & 45.4G & 84.2 \\ \hline
    \end{tabular}
    \label{tab:fc}
    \vspace{-3mm}
\end{table}

\vspace{2.8mm}
{\scriptsize
\bibliographystyle{_sty/ieee.bst}
\bibliography{_bib/Chenhui-Qudra}
}
\end{document}